\documentclass{article}

\usepackage[nonatbib,final]{nips_2017}

\usepackage{resizegather}
\usepackage{graphicx}
\usepackage[english]{babel}
\usepackage[utf8x]{inputenc}
\usepackage[T1]{fontenc}
\usepackage{fullpage}
\usepackage{bbold}
\def\decide#1#2{
  \mathrel{
    \mathop{
      \begin{array}{c}
        >\vspace{-1.4ex}\\<
      \end{array}
      }\limits_{#2}\limits^{#1}
    }
  }

\usepackage{amsmath,amssymb}
\usepackage{amsopn}
\usepackage{bbm}
\usepackage{amsthm}
\usepackage{graphicx}
\usepackage{subfigure}
\usepackage{multirow}

\usepackage[usenames,dvipsnames]{xcolor}
\usepackage[colorinlistoftodos]{todonotes}
\usepackage{algorithm}
\usepackage{algorithmic}

\DeclareMathOperator{\sgn}{sgn} 

\usepackage{wrapfig}
\usepackage{wrapfig,booktabs}

\title{Adaptive Classification for Prediction Under a Budget}

\author{
  Feng Nan \\
  Systems Engineering\\
  Boston University\\
  Boston, MA 02215\\
  \texttt{fnan@bu.edu} \\
	\And
	Venkatesh Saligrama \\
	Electrical Engineering\\
	Boston University \\
	Boston, MA 02215 \\
	\texttt{srv@bu.edu} \\
}

\begin{document}

\maketitle

\begin{abstract}
We propose a novel adaptive approximation approach for test-time resource-constrained prediction. Given an input instance at test-time, a gating function identifies a prediction model for the input among a collection of models. Our objective is to minimize overall average cost without sacrificing accuracy. We learn gating and prediction models on fully labeled training data by means of a bottom-up strategy. Our novel bottom-up method first trains a high-accuracy complex model. Then a low-complexity gating and prediction model are subsequently learnt to adaptively approximate the high-accuracy model in regions where low-cost models are capable of making highly accurate predictions. We pose an empirical loss minimization problem with cost constraints to jointly train gating and prediction models. On a number of benchmark datasets our method outperforms state-of-the-art achieving higher accuracy for the same cost.
\end{abstract}

\section{Introduction}
Resource costs arise during test-time prediction in a number of machine learning applications. 
Feature costs in Internet, Healthcare, and Surveillance applications arise due to 
to feature extraction time~\cite{xu2013cost}, and 
feature/sensor acquisition~\cite{trapeznikov:2013b}. 
%
The goal in such scenarios is to learn models on fully annotated training data that maintains high accuracy while meeting average resource constraints during prediction-time. 

There have been a number of promising 
approaches that focus on methods for reducing costs while improving overall accuracy ~\cite{Gao+Koller:NIPS11,DBLP:conf/icml/XuWC12,trapeznikov:2013b,Wang2014,ASTC_AAAI14,icml2015_nan15}. These methods are adaptive in that, at test-time, resources (features, computation etc) are allocated adaptively depending on the difficulty of the input. Many of these methods train models in a top-down manner, namely, attempt to build out the model by selectively adding the most cost-effective features to improve accuracy.

In contrast we propose a novel bottom-up approach. We train adaptive models on annotated training data by selectively identifying parts of the input space for which high accuracy can be maintained at a lower cost.  
The principle advantage of our method is twofold. First, our approach can be readily applied to cases where it is desirable to reduce costs of an existing high-cost legacy system. Second, training top-down models leads to fundamental combinatorial issues in multi-stage search over all feature subsets (see Sec.~\ref{sec:related}). In contrast, we bypass many of these issues by posing a natural adaptive approximation objective to partition the input space into easy and hard cases. Our key insight is that reducing costs of an existing high-accuracy system (bottom-up approach) is generally easier by selectively identifying redundancies using L1 or other group sparse norms. 

%

\begin{wrapfigure}{r}{0.5\textwidth}
\includegraphics[width=0.5\textwidth]{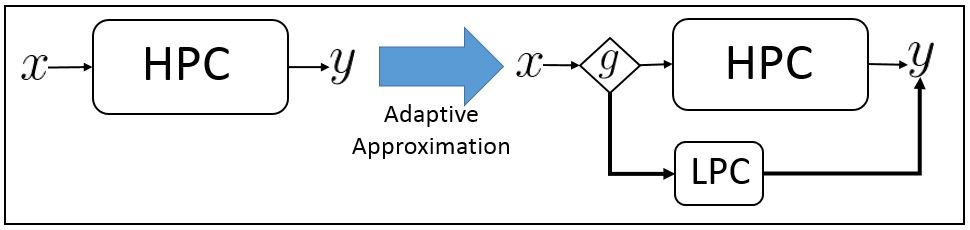}
\caption{Single stage schematic of our approach. We learn low-cost gating $g$ and a low-prediction cost (LPC) model to adaptively approximate a high prediction cost (HPC) model.} \label{fig:adaptive_approximation}
\vspace{-.5cm}
\end{wrapfigure}

In particular, when no legacy system is available, our method consists of first learning a high-accuracy model that minimizes the empirical loss regardless of costs. The resulting high prediction-cost model (HPC) can be readily trained using any of the existing methods. Next, we then jointly learn a low-cost gating function as well as a low prediction-cost (LPC) model so as to \emph{adaptively approximate} the high-accuracy model by identifying regions of input space where a low-cost gating and LPC model are adequate to achieve high-accuracy. At test-time, for each input instance, the gating function decides whether or not the LPC model is adequate for accurate classification. Intuitively, ``easy'' examples can be correctly classified using only an LPC model while ``hard'' examples require HPC model. By identifying which of the input instances can be classified accurately with LPCs we bypass the utilization of HPC model, thus reducing average prediction cost. Figure \ref{fig:adaptive_approximation} is a schematic of our approach, where $x$ is feature vector and $y$ is the predicted label; we aim to learn $g$ and an LPC model to adaptively approximate the HPC. 

The problem would be simpler if our task were to primarily partition the input space into regions where LPC models would suffice. The difficulty is that we must also learn a low gating-cost function capable of identifying input instances for which LPC suffices. 
Since both prediction and gating account for cost, we favor design strategies that lead to shared features and decision architectures between the gating function and the LPC model. 
We pose the problem as a discriminative empirical risk minimization problem that jointly optimizes for gating and prediction models in terms of a joint margin-based objective function.
The resulting objective is separately convex in gating and prediction functions. We propose an alternating minimization scheme that is guaranteed to converge since with appropriate choice of loss-functions (for instance, logistic loss), each optimization step amounts to a probabilistic approximation/projection (I-projection/M-projection) onto a probability space. While our method can be recursively applied in multiple stages to successively approximate the adaptive system obtained in the previous stage, thereby refining accuracy-cost trade-off, we observe that on benchmark datasets even a single stage of our method outperforms state-of-art in accuracy-cost performance.

\section{Related Work} \label{sec:related}
Learning decision rules to minimize error subject to a budget constraint during prediction-time is an area of active interest
\cite{Gao+Koller:NIPS11,DBLP:conf/icml/XuWC12,trapeznikov:2013b,weiss_taskar2013,Wang2014, NIPS2015_5982,ASTC_AAAI14,NanNIPS2016}.

{\it Pre-trained Models:} In one instantiation of these methods it is assumed that there exists a collection of prediction models with amortized costs~\cite{weiss_taskar2013,trapeznikov:2013b} so that a natural ordering of prediction models can be imposed. In other instances, the feature dimension is assumed to be sufficiently low so as to admit an exhaustive enumeration of all the combinatorial possibilities ~\cite{Wang2014,NIPS2015_5982}. These methods then learn a policy to choose amongst the ordered prediction models. In contrast we do not impose any of these restrictions. 

\noindent
{\it Top-Down Methods:}
For high-dimensional spaces, many existing approaches focus on learning complex adaptive decision functions top-down \cite{Gao+Koller:NIPS11,DBLP:conf/icml/XuWC12,ASTC_AAAI14,NIPS2015_5982}. Conceptually, during training, top-down methods acquire new features based on their utility value. This requires exploration of partitions of the input space together with different combinatorial low-cost feature subsets that would result in higher accuracy. These methods are based on multi-stage exploration leading to combinatorially hard problems. Different novel relaxations and greedy heuristics have been developed in this context. 

{\it Bottom-up Methods:}
Our work is somewhat related to \cite{NanNIPS2016}, who propose to prune a fully trained random forests (RF) to reduce costs. Nevertheless, in contrast to our adaptive system, their perspective is to compress the original model and utilize the pruned forest as a stand-alone model for test-time prediction. Furthermore, their method is specifically tailored to random forests. 

Another set of related work includes classifier cascade \cite{DBLP:journals/jmlr/ChenXWCK12} and decision DAG \cite{DBLP:conf/icml/Busa-FeketeBK12}, both of which aim to re-weight/re-order a set of pre-trained base learners to reduce prediction budget. Our method, on the other hand, only requires to pre-train a high-accuracy model and jointly learns the low-cost models to approximate it; therefore ours can be viewed as complementary to the existing work.
The teacher-student framework \cite{LopSchBotVap16} is also related to our bottom-up approach; a low-cost student model learns to approximate the teacher model so as to meet test-time budget. However, the goal there is to learn a better stand-alone student model. In contrast, we make use of both the low-cost (student) and high-accuracy (teacher) model during prediction via a gating function, which learns the limitation of the low-cost (student) model and consult the high-accuracy (teacher) model if necessary, thereby avoiding accuracy loss. 

Our composite system is also related to HME~\cite{Jordan:1994:HME:188104.188106}, which learns the composite system based on max-likelihood estimation of models. A major difference is that HME does not address budget constraints. A fundamental aspect of budget constraints is the resulting asymmetry, whereby, we start with an HPC model and sequentially approximate with LPCs. This asymmetry leads us to propose a bottom-up strategy where the high-accuracy predictor can be separately estimated and is critical to posing a direct empirical loss minimization problem.

\section{Problem Setup}
We consider the standard learning scenario of resource constrained prediction with feature costs. A training sample $S=\{(x^{(i)},y^{(i)}):{i=1,\dots,N}\}$ is generated i.i.d. from an unknown distribution,  where $x^{(i)} \in \Re^K$ is the feature vector with an acquisition cost $c_\alpha\geq 0$ assigned to each of the features $\alpha=1,\dots,K$ and $y^{(i)}$ is the label for the $i^{\mbox{th}}$ example. In the case of multi-class classification $y \in \{1,\dots,M\}$, where $M$ is the number of classes. Let us consider a single stage of our training method in order to formalize our setup. The model, $f_0$, is a high prediction-cost (HPC) model, which is either a priori known, or which we train to high-accuracy regardless of cost considerations. We would like to learn an alternative low prediction-cost (LPC) model $f_1$. 
Given an example $x$, at test-time, we have the option of selecting which model, $f_0$ or $f_1$, to utilize to make a prediction. 
The accuracy of a prediction model $f_z$ is modeled by a loss function $\ell(f_z(x),y),z\in \{0,1\}$. We exclusively employ the logistic loss function in  binary classification: $\ell(f_z(x),y)=\log(1+\exp(-yf_z(x))$, although our framework allows other loss models. For a given $x$, we assume that once it pays the cost to acquire a feature, its value can be efficiently cached; its subsequent use does not incur additional cost. Thus, the cost of utilizing a particular prediction model, denoted by $c(f_z,x)$, is computed as the sum of the acquisition cost of \emph{unique} features required by $f_z$.

\noindent
{\bf Oracle Gating:} Consider a general gating likelihood function $q(z|x)$ with $z \in \{0,\,1\}$, that outputs the likelihood of sending the input $x$ to  a prediction model, $f_z$.  
 The overall empirical loss is:
\begin{align*}
\mathbb{E}_{S_n} \mathbb{E}_{q(z | x)} [\ell(f_z(x),y)] =\mathbb{E}_{S_n} [\ell(f_0(x),y)]+ \mathbb{E}_{S_n} \big[q(1 | x) \underbrace{(\ell(f_1(x),y) - \ell(f_0(x),y))\big]}_{\textcolor{red}{Excess\, Loss}}
\end{align*}
The first term only depends on $f_0$, and from our perspective a constant. 
Similar to average loss we can write the average cost as (assuming gating cost is negligible for now):
\begin{align*}
\mathbb{E}_{S_n}\mathbb{E}_{q(z|x)} [c(f_z,x)] = \mathbb{E}_{S_n} [c(f_0,x)]-\mathbb{E}_{S_n} [q(1|x)\underbrace{(c(f_0,x)-c(f_1,x))}_{\textcolor{red}{Cost\, Reduction}}],
\end{align*}
where the first term is again constant. We can characterize the optimal gating function (see \cite{trapeznikov:2013b}) that minimizes the overall average loss subject to average cost constraint:
$$
\overbrace{\ell(f_1,x)-\ell(f_0,x)}^{\textcolor{red}{Excess\, loss}} \decide{q(1|x)=0}{q(1|x)=1} \eta \overbrace{(c(f_0,x)-c(f_1,x))}^{\textcolor{red}{Cost\, reduction}} 
$$
for a suitable choice $\eta \in \mathbb{R}$. This characterization encodes the important principle that if the marginal cost reduction is smaller than the excess loss, we opt for the HPC model. 
Nevertheless, this characterization is generally infeasible. Note that the LHS depends on knowing how well HPC performs on the input instance. Since this information is unavailable, this target can be unreachable with low-cost gating. 

\noindent
{\bf Gating Approximation:}
Rather than directly enforcing a low-cost structure on $q$, we decouple the constraint and introduce a parameterized family of gating functions $g \in {\cal G}$ that attempts to mimic (or approximate) $q$. To ensure such approximation, we can minimize some distance measure $D(q(\cdot |x),g(x))$. 
A natural choice for an approximation metric is the Kullback-Leibler (KL) divergence although other choices are possible. The KL divergence between $q$ and $g$ is given by $D_{KL}(q(\cdot |x)\|g(x)) = \sum_z q(z |x) \log(q(z |x)/\sigma(\sgn(0.5-z)g(x)))$, where $\sigma(s)=1/(1+e^{-s})$ is the sigmoid function. Besides KL divergence, we have also proposed another symmetrized metric fitting $g$ directly to the log odds ratio of $q$. See Suppl. Material for details.

\noindent
{\bf Budget Constraint:}
With the gating function $g$, the cost of predicting $x$ depends on whether the example is sent to $f_0$ or $f_1$. Let $c(f_0, g,x)$ denote the feature cost of passing $x$ to $f_0$ through $g$. As discussed, this is equal to the sum of the acquisition cost of unique features required by $f_0$ and $g$ for $x$. Similarly $c(f_1, g,x)$ denotes the cost if $x$ is sent to $f_1$ through $g$. In many cases the cost $c(f_z,g,x)$ is independent of the example $x$ and depends primarily on the model being used. This is true for linear models where each $x$ must be processed through the same collection of features. For these cases $c(f_z,g,x) \triangleq c(f_z,g)$. The total budget simplifies to: $\mathbb{E}_{S_n}[q(0|x)]c(f_0,g)+(1-\mathbb{E}_{S_n}[q(0|x)])c(f_1,g)=c(f_1,g)+\mathbb{E}_{S_n}[q(0|x)] (c(f_0,g)-c(f_1,g))$.
The budget thus depends on 3 quantities: $\mathbb{E}_{S_n}[q(0|x)]$, $c(f_1,g)$ and $c(f_0,g)$. Often $f_0$ is a high-cost model that requires most, if not all, of features so $c(f_0,g)$ can be considered a large constant.

Thus, to meet the budget constraint, we would like to have (a) low-cost $g$ and $f_1$ (small $c(f_1,g)$); and (b) small fraction of examples being sent to the high-accuracy model (small $\mathbb{E}_{S_n}[q(0|x)]$).
We can therefore split the budget constraint into two separate objectives: (a) ensure low-cost through penalty $\Omega(f_1,g)=\gamma \sum_{\alpha}c_\alpha \|V_{\alpha}+W_{\alpha}\|_0$, where $\gamma$ is a tradeoff parameter and the indicator variables $V_{\alpha},W_{\alpha} \in \{0,1\}$ denote whether or not the feature $\alpha$ is required by $f_1$ and $g$, respectively. Depending on the model parameterization, we can approximate $\Omega(f_1,g)$ using a group-sparse norm or in a stage-wise manner as we will see in Algorithms~\ref{alg:Adapt-lin} and \ref{alg:Adapt-gbrt}.  
(b)  Ensure only $\text{P}_{\text{full}}$ fraction of examples are sent to $f_0$ via the constraint $\mathbb{E}_{S_n}[q(0|x)] \leq P_{\text{full}}$.

\noindent
{\bf Putting Together:} We are now ready to pose our general optimization problem:
\begin{align*} \tag{OPT}\label{eq:OPT}
\hspace{-.15cm}
\min_{f_1 \in {\cal F}, g \in {\cal G}, q}
&\mathbb{E}_{S_n}\overbrace{\sum\limits_{z}[ q(z|x)\ell(f_z(x),y)]}^{\textcolor{red}{Losses}} + \overbrace{D(q(\cdot|x),g(x))}^{\textcolor{red}{Gating\, Approx}} +\overbrace{\Omega(f_1,g)}^{\textcolor{red}{Feature\, Costs}}\\
\text{subject to: } & \mathbb{E}_{S_n}[q(0|x)] \leq P_{\text{full}}. \,\,\,   \textcolor{red}{(Fraction\, to \,} \color{red}{f_0} \textcolor{red}{)}   
\end{align*}
The objective function penalizes excess loss and ensures through the second term that this excess loss can be enforced through admissible gating functions. The third term penalizes the feature cost usage of $f_1$ and $g$. The budget constraint limits the fraction of examples sent to the costly model $f_0$.

{\noindent
{\emph{Remark 1}:}
}
We presented the case for a single stage approximation system. However, it is straightforward to recursively continue this process. We can then view the composite system $f_0 \triangleq (g,f_1, f_0)$ as a black-box predictor and train a new pair of gating and prediction models to approximate the composite system. 

{\noindent
{\emph{Remark 2}:}
}
To limit the scope of our paper, we focus on reducing feature acquisition cost during prediction as it is a more challenging (combinatorial) problem. However, other prediction-time costs such as computation cost can be encoded in the choice of functional classes $\cal F$ and $\cal G$ in \eqref{eq:OPT}. 

{\noindent
{\emph{Surrogate Upper Bound of Composite System}:}
}
We can get better insight for the first two terms of the objective in \eqref{eq:OPT} if we view $z \in \{0,1\}$ as a latent variable and consider the composite system $\Pr(y|x) = \sum_z \Pr(z|x;g) \Pr(y|x,f_z)$. A standard application of Jensen's inequality reveals that, 
$-\log(\Pr(y|x)) \leq \mathbb{E}_{q(z|x)} \ell(f_z(x),y) + D_{KL}(q(z|x)\|\Pr(z|x;g))$. Therefore, the conditional-entropy of the composite system is bounded by the expected value of our loss function (we overload notation and represent random-variables in lower-case format):
$$
H(y \mid x) \triangleq \mathbb{E}[-\log(\Pr(y|x))] \leq \mathbb{E}_{x\times y}[\mathbb{E}_{q(z|x)} \ell(f_z(x),y) + D_{KL}(q(z|x)\|\Pr(z|x;g))].
$$
This implies that the first two terms of our objective attempt to bound the loss of the composite system; the third term in the objective together with the constraint serve to enforce budget limits on the composite system. 

\noindent
{\it Group Sparsity:} Since the cost for feature re-use is zero we encourage feature re-use among gating and prediction models. So the fundamental question here is:
{\emph{How to choose a common, sparse (low-cost) subset of features on which both $g$ and $f_1$ operate, such that $g$ can effective gate examples between $f_1$ and $f_0$ for accurate prediction?}}
This is a hard combinatorial problem. The main contribution of our paper is to address it using the general optimization framework of \eqref{eq:OPT}.

\section{Algorithms}
To be concrete, we instantiate our general framework \eqref{eq:OPT} into two algorithms via different parameterizations of $g,f_1$: \textsc{Adapt-lin} for the linear class and \textsc{Adapt-Gbrt} for the non-parametric class. Both of them use the KL-divergence as distance measure. We also provide a third algorithm \textsc{Adapt-Lstsq} that uses the symmetrized distance in the Suppl. Material. All of the algorithms perform alternating minimization of \eqref{eq:OPT} over $q,g,f_1$. 
Note that convergence of alternating minimization follows as in \cite{NIPS2007_3170}. Common to all of our algorithms, we use two parameters to control cost: $\text{P}_{\text{full}}$ and $\gamma$. In practice they are swept to generate various cost-accuracy tradeoffs and we choose the best one satisfying the budget $B$ using validation data.

\noindent
{\bf \textsc{Adapt-lin}:} Let $g(x) = g^Tx$ and $f_1(x)=f_1^Tx$ be linear classifiers. A feature is used if the corresponding component is non-zero: $V_\alpha=1$ if $f_{1,\alpha} \neq 0$, and $W_\alpha=1$ if $g_\alpha \neq 0$. 
\begin{wrapfigure}{r}{0.5\textwidth}
\vspace{-0.55cm}
\begin{minipage}{.49\textwidth}
\begin{algorithm}[H]
   \caption{\textsc{Adapt-Lin}}
   \label{alg:Adapt-lin}
\begin{algorithmic}
   \STATE {\bfseries Input:} $(x^{(i)},y^{(i)}),\text{P}_{\text{full}}, \gamma$
   \STATE Train $f_0$. Initialize $g, f_1$.
   \REPEAT
   \STATE Solve (OPT1) for $q$ given $g, f_1$. 
   \STATE Solve (OPT2) for $g, f_1$ given $q$.
   \UNTIL{convergence}
\end{algorithmic}
\end{algorithm}
\vspace{-.70cm}
\begin{algorithm}[H]
   \caption{\textsc{Adapt-Gbrt}}
   \label{alg:Adapt-gbrt}
\begin{algorithmic}
   \STATE {\bfseries Input:} $(x^{(i)},y^{(i)}),\text{P}_{\text{full}}, \gamma$
   \STATE Train $f_0$. Initialize $g,f_1$.
   \REPEAT
   \STATE Solve (OPT1) for $q$ given $g, f_1$. 
   \FOR{$t=1$ {\bfseries to} $T$}
   		\STATE Find $f_1^t$ using CART to minimize \eqref{eq:impurity_h}.
   		\STATE $f_1=f_1 + f_1^t$.
   		\STATE For each feature $\alpha$ used, set $u_\alpha=0$.
   		\STATE Find $g^t$ using CART to minimize \eqref{eq:impurity_g}.
   		\STATE $g=g+ g^t$.
   		\STATE For each feature $\alpha$ used, set $u_\alpha=0$.
   \ENDFOR
   \UNTIL{convergence}
\end{algorithmic}
\end{algorithm}
\vspace{-.80cm}
\end{minipage}%
\vspace{-.90cm}
\end{wrapfigure}
%
The minimization for $q$ solves the following problem:
\begin{equation}
\begin{array}{rlll}\tag{OPT1}\label{eq:OPT1}
\displaystyle \min_{q} &  \multicolumn{2}{l}{\frac{1}{N} \sum_{i=1}^{N} \left [(1-q_i)A_i+q_iB_i - H(q_i)\right ]} \\
\textrm{s.t.} &  \frac{1}{N} \sum_{i=1}^{N} q_i \leq \text{P}_{\text{full}}, \end{array}
\end{equation}
where we have used shorthand notations $q_i = q(z=0|x^{(i)})$, $H(q_i)=-q_i\log(q_i)-(1-q_i)\log(1-q_i)$, $A_i=\log(1+e^{-y^{(i)}f_1^Tx^{(i)}})+\log(1+e^{g^Tx^{(i)}})$ and $B_i=-\log p(y^{(i)}|z^{(i)}=0;f_0)+\log(1+e^{-g^Tx^{(i)}})$.
This optimization has a closed form solution: $q_i = 1/(1+e^{B_i-A_i+\beta})$ for some non-negative constant $\beta$ such that the constraint is satisfied. This optimization is also known as I-Projection in information geometry because of the entropy term \cite{NIPS2007_3170}.
Having optimized $q$, we hold it constant and minimize with respect to $g,f_1$ by solving the problem \eqref{eq:OPT2}, where we have relaxed the non-convex cost $\sum_{\alpha}c_\alpha \|V_{\alpha}+W_{\alpha}\|_0$ into a $L_{2,1}$ norm for group sparsity and a tradeoff parameter $\gamma$ to make sure the feature budget is satisfied. Once we solve for $g,f_1$, we can hold them constant and minimize with respect to $q$ again. \textsc{Adapt-Lin} is summarized in Algorithm~\ref{alg:Adapt-lin}. 
\begin{equation}\tag{OPT2}\label{eq:OPT2}
\displaystyle \min_{g,f_1} \frac{1}{N} \sum_{i=1}^{N} \left [(1-q_i) \left (\log(1+e^{-y^{(i)}f_1^Tx^{(i)}})+\log(1+e^{g^Tx^{(i)}}) \right )+q_i \log(1+e^{-g^Tx^{(i)}})\right ] + \gamma \sum_{\alpha} \sqrt{g_{\alpha}^2+f_{1,\alpha}^2}.
\end{equation}

\noindent
{\bf \textsc{Adapt-Gbrt}:} We can also consider the non-parametric family of classifiers such as gradient boosted trees \cite{Friedman00greedyfunction}: $g(x) = \sum_{t=1}^{T} g^t(x)$ and $f_1(x)=\sum_{t=1}^{T} f_1^t(x)$, where $g^t$ and $f_1^t$ are limited-depth regression trees.
Since the trees are limited to low depth, we assume that the feature utility of each tree is example-independent: $V_{\alpha,t}(x)\approxeq V_{\alpha,t}, W_{\alpha,t}(x)\approxeq W_{\alpha,t},\forall x$. $V_{\alpha,t}=1$ if feature $\alpha$ appears in $f_1^t$, otherwise $V_{\alpha,t}=0$, similarly for $W_{\alpha,t}$.
The optimization over $q$ still solves \eqref{eq:OPT1}. We modify $A_i=\log(1+e^{-y^{(i)}f_1(x^{(i)})})+\log(1+e^{g(x^{(i)})})$ and $B_i=-\log p(y^{(i)}|z^{(i)}=0;f_0)+\log(1+e^{-g(x^{(i)})})$. Next, to minimize over $g,f_1$, denote loss: 
\begin{align*}
\ell(f_1,g) =\frac{1}{N} \sum_{i=1}^{N} \Bigg[(1-q_i)\cdot 
\left (\log(1+e^{-y^{(i)}f_1(x^{(i)})})+\log(1+e^{g(x^{(i)})}) \right )+q_i \log(1+e^{-g(x^{(i)})})\Bigg],
\end{align*}
which is essentially the same as the first part of the objective in \eqref{eq:OPT2}. 
Thus, we need to minimize $\ell(f_1,g)+\Omega(f_1,g)$ with respect to $f_1$ and $g$. Since both $f_1$ and $g$ are gradient boosted trees, we naturally adopt a stage-wise approximation for the objective. In particular, we define an impurity function which on the one hand approximates the negative gradient of $\ell(f_1,g)$ with the squared loss, and on the other hand penalizes the initial acquisition of features by their cost $c_\alpha$. 
To capture the initial acquisition penalty, we let $u_\alpha \in \{0,1\}$ indicates if feature $\alpha$ has already been used in previous trees ($u_\alpha=0$), or not ($u_\alpha=1$). $u_\alpha$ is updated after adding each tree. Thus we arrive at the following impurity for $f_1$ and $g$, respectively:
\begin{equation} \label{eq:impurity_h}
\frac{1}{2}\sum_{i=1}^{N}(-\frac{\partial \ell(f_1,g)}{\partial f_1(x^{(i)})} - f_1^t(x^{(i)}))^2+\gamma\sum_{\alpha} u_\alpha c_\alpha V_{\alpha,t},
\end{equation}
\begin{equation} \label{eq:impurity_g}
\frac{1}{2}\sum_{i=1}^{N}(-\frac{\partial \ell(f_1,g)}{\partial g(x^{(i)})} - g^t(x^{(i)}))^2+\gamma\sum_{\alpha} u_\alpha c_\alpha W_{\alpha,t}.
\end{equation}
Minimizing such impurity functions balances the need to minimize loss and re-using the already acquired features. Classification and Regression Tree (CART) \cite{breiman1984classification} can be used to construct decision trees with such an impurity function. \textsc{Adapt-GBRT} is summarized in Algorithm~\ref{alg:Adapt-gbrt}. 
Note that a similar impurity is used in \textsc{GreedyMiser} \cite{DBLP:conf/icml/XuWC12}.
Interestingly, if $\text{P}_{\text{full}}$ is set to 0, all the examples are forced to $f_1$, then \textsc{Adapt-Gbrt} exactly recovers the \textsc{GreedyMiser}. In this sense, \textsc{GreedyMiser} is a special case of our algorithm. As we will see in the next section, thanks to the bottom-up approach, \textsc{Adapt-Gbrt} benefits from high-accuracy initialization and is able to perform accuracy-cost tradeoff in accuracy levels beyond what is possible for \textsc{GreedyMiser}.

\vspace{-0.15in}
\section{Experiments}
\noindent
{\bf \textsc{Baseline Algorithms}:} We consider the following simple L1 baseline approach for learning $f_1$ and $g$: first perform a L1-regularized logistic regression on all data to identify a relevant, sparse subset of features; then learn $f_1$ using training data restricted to the identified feature(s); finally, learn $g$ based on the correctness of $f_1$ predictions as pseudo labels (i.e. assign pseudo label 1 to example $x$ if $f_1(x)$ agrees with the true label $y$ and 0 otherwise). We also compare with two state-of-the-art feature-budgeted algorithms: \textsc{GreedyMiser}\cite{DBLP:conf/icml/XuWC12} - a top-down method that builds out an ensemble of gradient boosted trees with feature cost budget; 
and \textsc{BudgetPrune}\cite{NanNIPS2016} - a bottom-up method that prunes a random forest with feature cost budget. A number of other methods such as ASTC \cite{ASTC_AAAI14} and CSTC \cite{xu2013cost} are omitted as they have been shown to under-perform \textsc{GreedyMiser} on the same set of datasets \cite{icml2015_nan15}. Detailed experiment setups can be found in the Suppl. Material.

We first visualize/verify the adaptive approximation ability of \textsc{Adapt-Lin} and \textsc{Adapt-Gbrt} on the Synthetic-1 dataset without feature costs. Next, we illustrate the key difference between \textsc{Adapt-Lin} and the L1 baseline approach on the Synthetic-2 as well as the Letters datasets. Finally, we compare \textsc{Adapt-Gbrt} with state-of-the-art methods on several resource constraint benchmark datasets.   
\begin{figure*}[htb!]
\centering
\subfigure[Input Data]{\includegraphics[width=.3\linewidth,height=.18\linewidth]{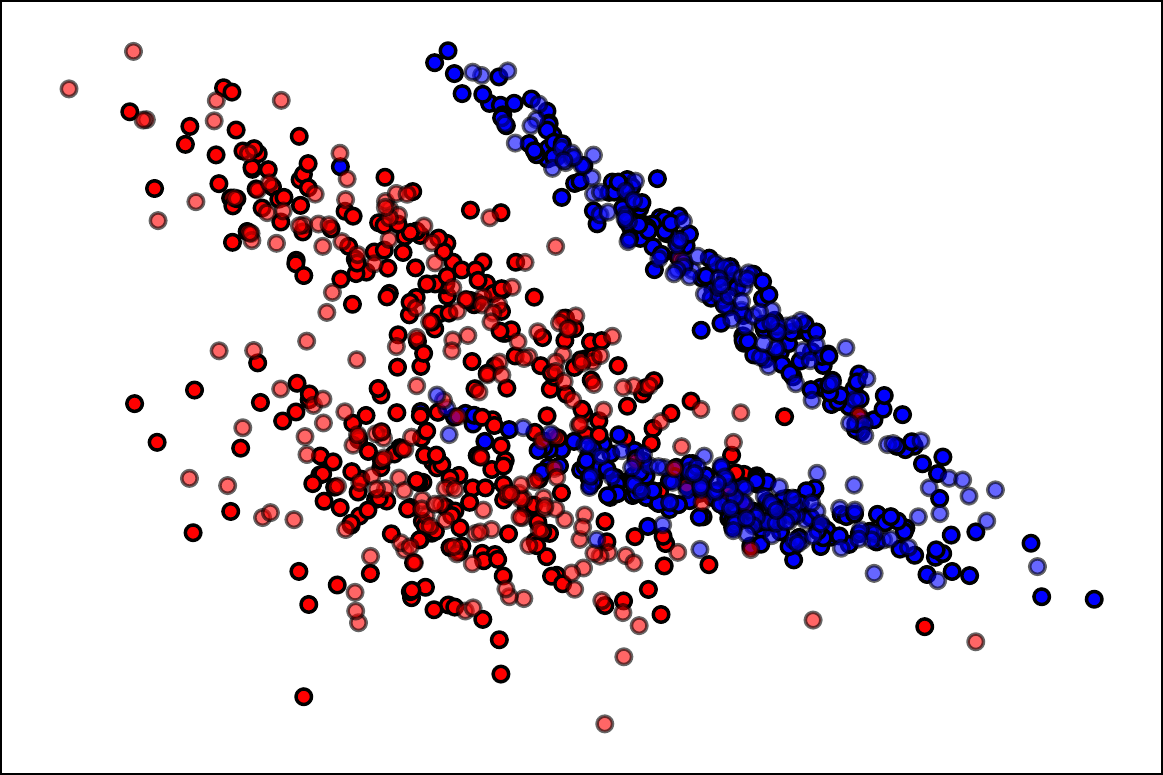}}
\vspace{-0.1in}
\subfigure[Lin Initialization]{\includegraphics[width=.3\linewidth,height=.18\linewidth]{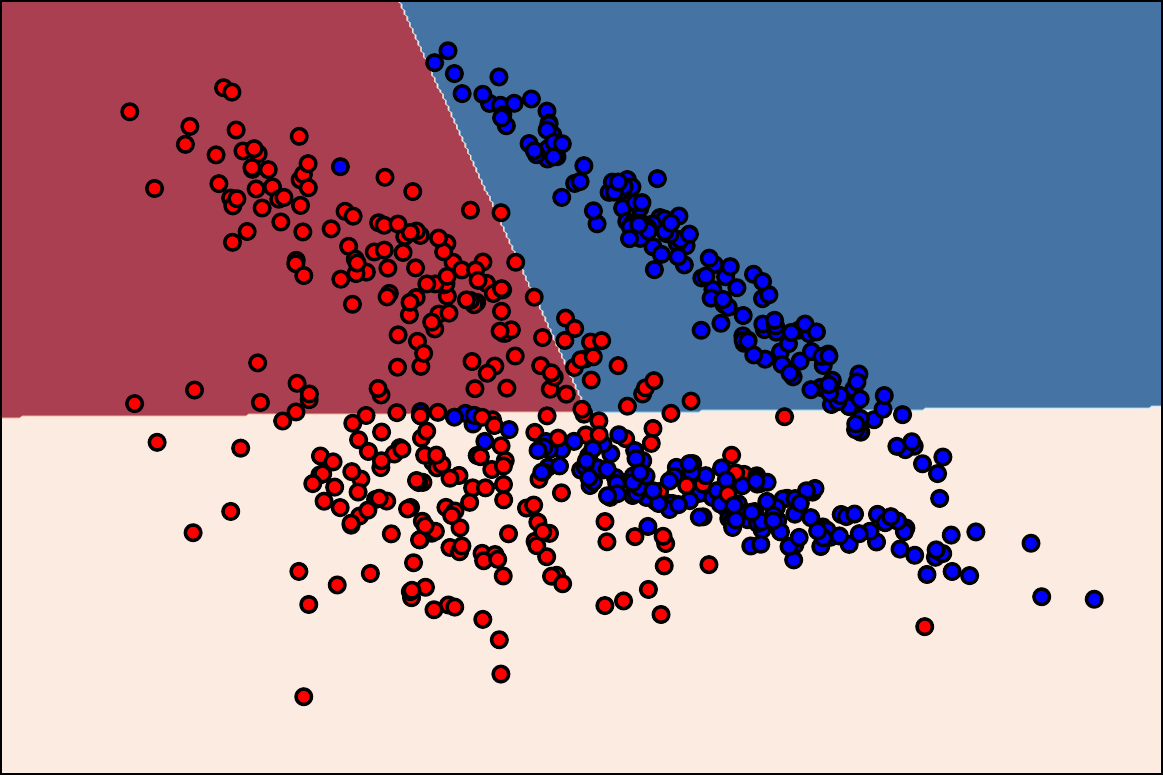}}
\subfigure[Lin after 10 iterations]{\includegraphics[width=.3\linewidth,height=.18\linewidth]{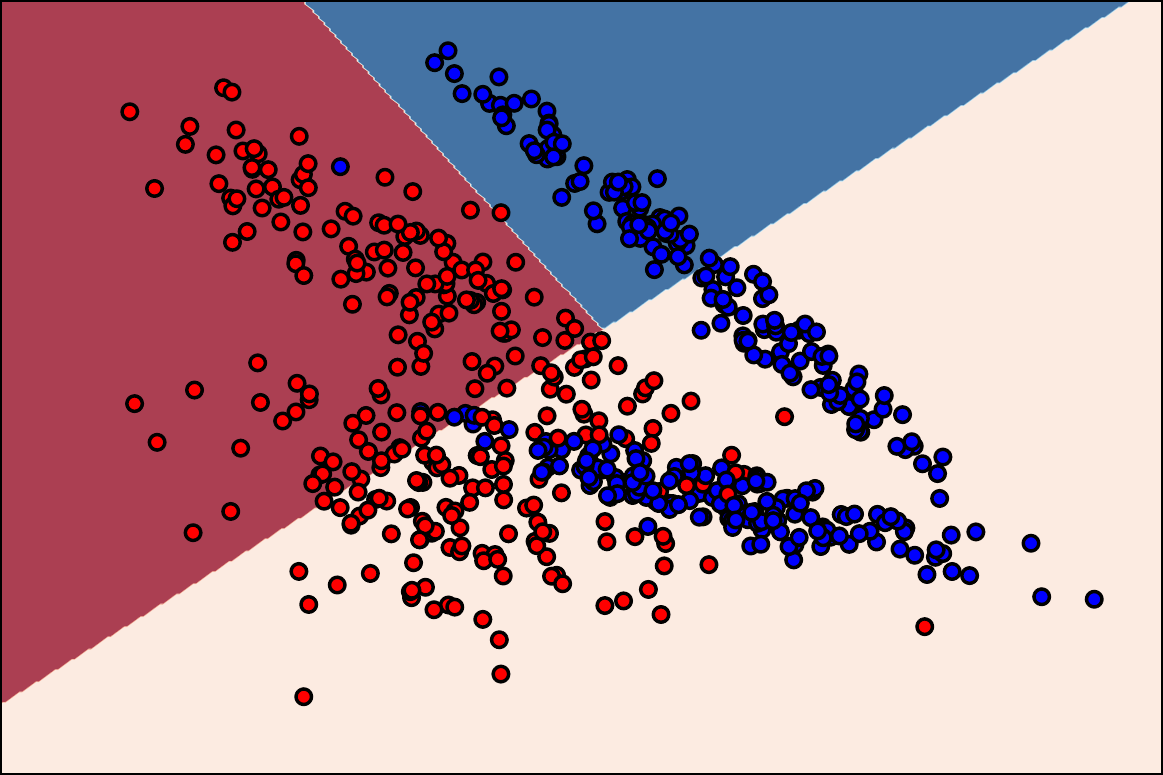}}
\\
\subfigure[RBF Contour]{\includegraphics[width=.3\linewidth,height=.18\linewidth]{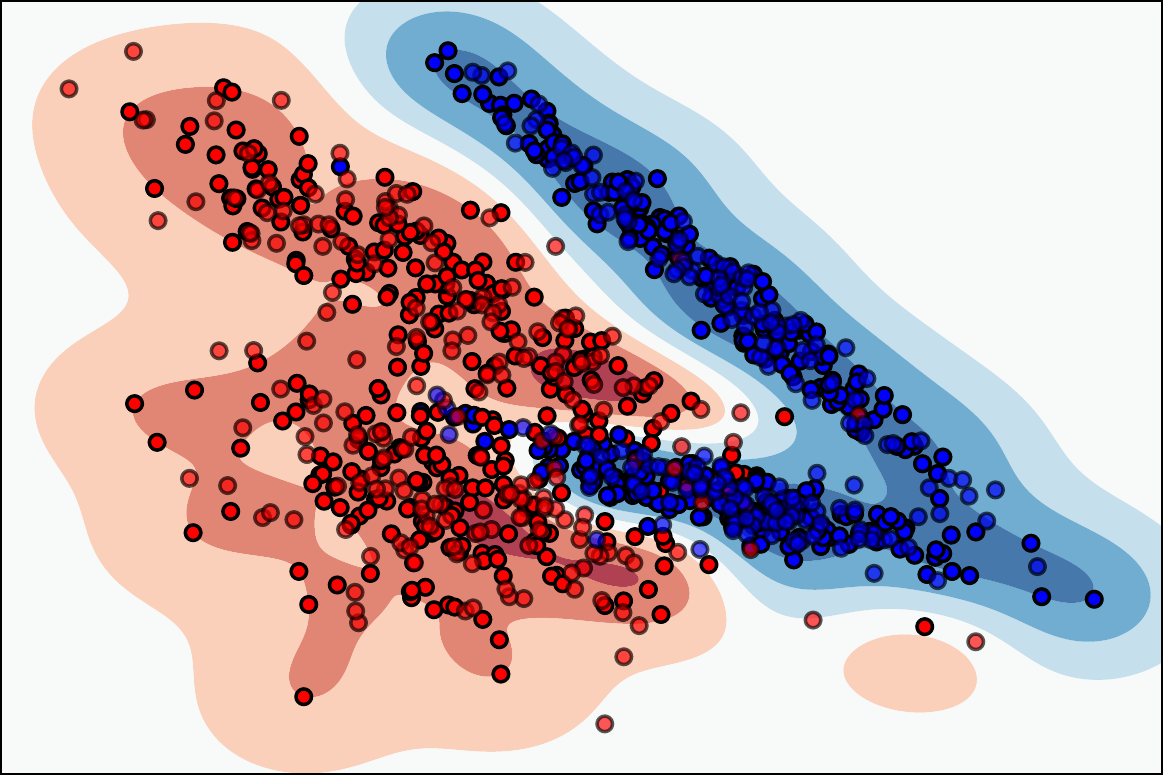}}
\subfigure[Gbrt Initialization]{\includegraphics[width=.3\linewidth,height=.18\linewidth]{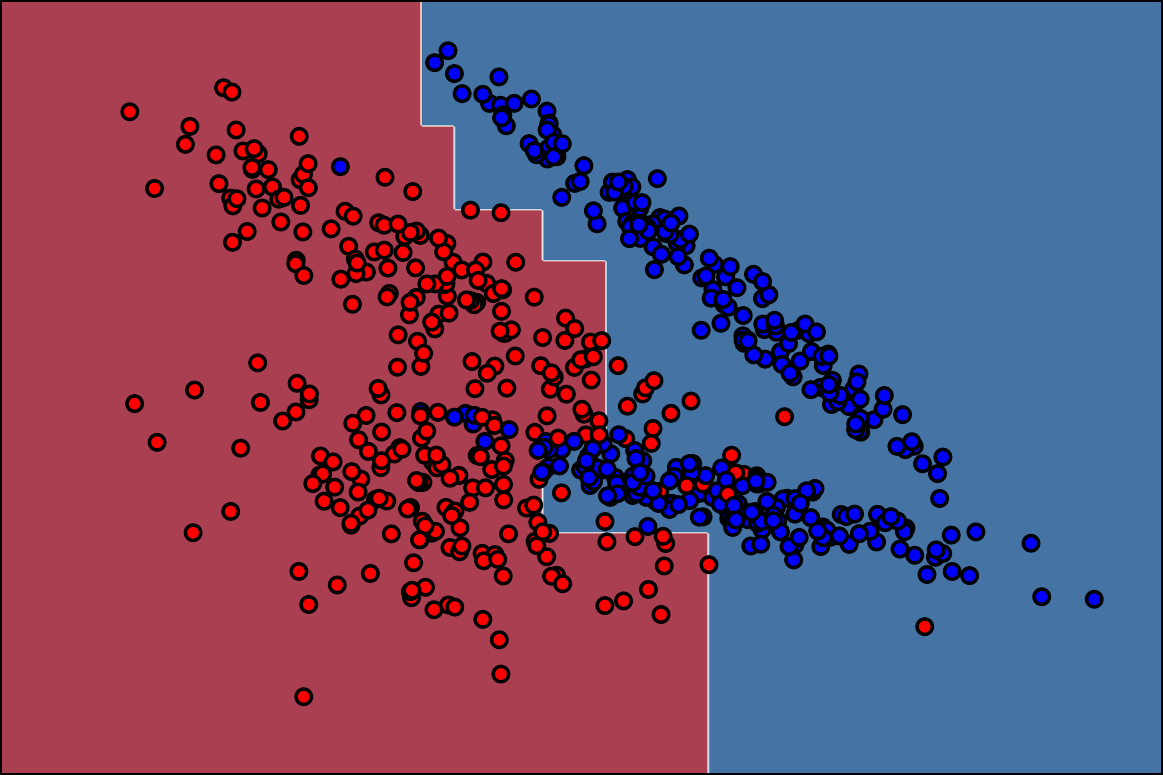}}
\subfigure[Gbrt after 10 iterations]{\includegraphics[width=.3\linewidth,height=.18\linewidth]{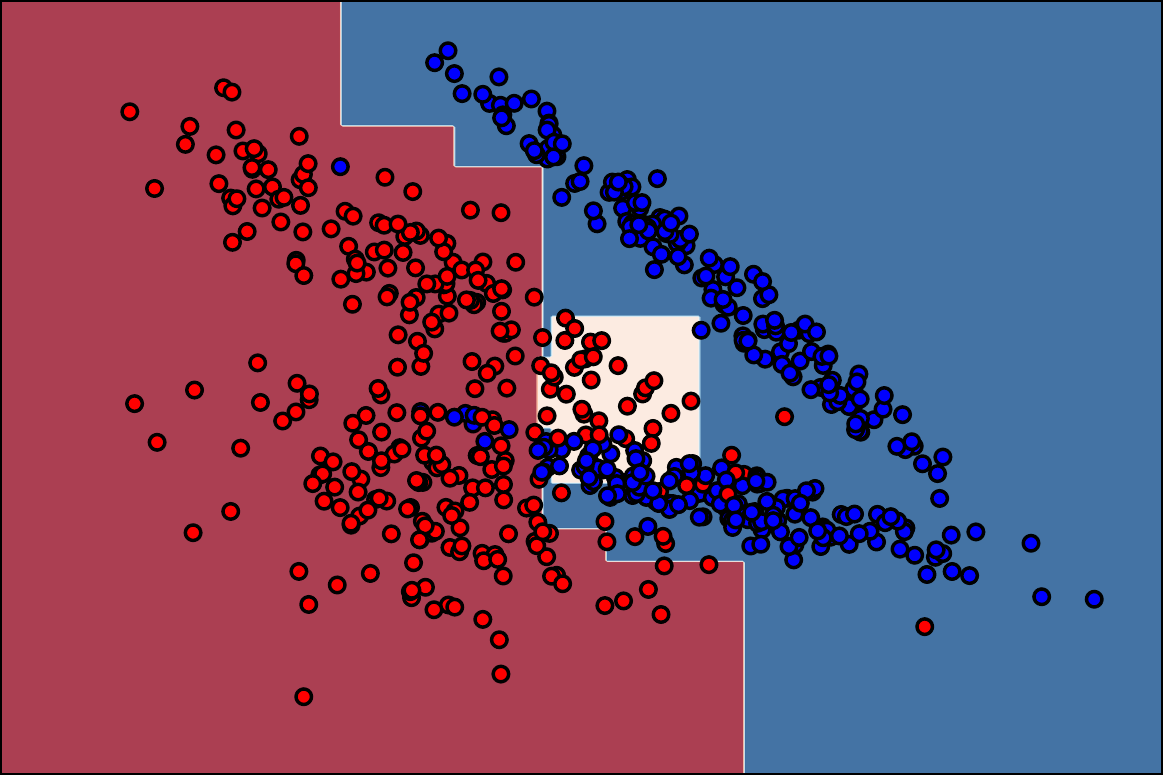}}
\caption{Synthetic-1 experiment without feature cost. (a): input data. (d): decision contour of RBF-SVM as $f_0$. (b) and (c): decision boundaries of linear $g$ and $f_1$ at initialization and after 10 iterations of \textsc{Adapt-Lin}. (e) and (f): decision boundaries of boosted tree $g$ and $f_1$ at initialization and after 10 iterations of \textsc{Adapt-Gbrt}. Examples in the beige areas are sent to $f_0$ by the $g$.}
\label{fig:synthetic_lin_xgb}
\vspace{-0.1in}
\end{figure*}

\noindent
{\bf \textsc{Power of Adaptation:}} We construct a 2D binary classification dataset (Synthetic-1) as shown in (a) of Figure \ref{fig:synthetic_lin_xgb}. We learn an RBF-SVM as the high-accuracy classifier $f_0$ as in (d). 
To better visualize the adaptive approximation process in 2D, we turn off the feature costs (i.e. set $\Omega(f_1,g)$ to 0 in \eqref{eq:OPT}) and run \textsc{Adapt-Lin} and \textsc{Adapt-Gbrt}. The initializations of $g$ and $f_1$ in (b) results in wrong predictions for many red points in the blue region. After 10 iterations of \textsc{Adapt-Lin}, $f_1$ adapts much better to the local region assigned by $g$ while $g$ sends about 60\% ($P_\text{full}$) of examples to $f_0$. Similarly, the initialization in (e) results in wrong predictions in the blue region. \textsc{Adapt-Gbrt} is able to identify the ambiguous region in the center and send those examples to $f_0$ via $g$. Both of our algorithms maintain the same level of prediction accuracy as $f_0$ yet are able to classify large fractions of examples via much simpler models. 
\begin{wrapfigure}{r}{0.4\textwidth}
\includegraphics[width=0.4\textwidth]{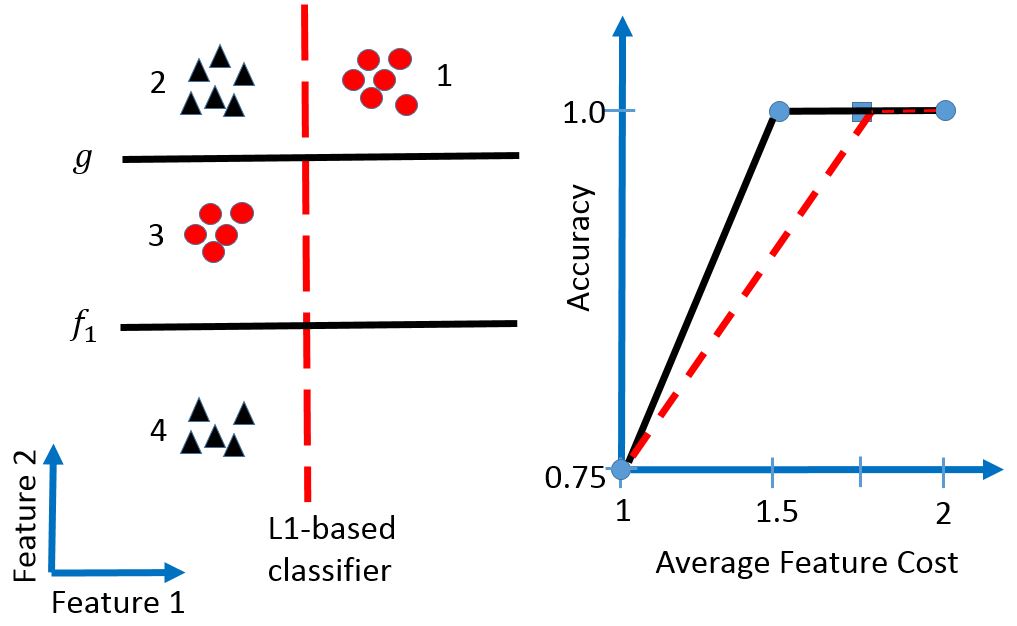}
\vspace{-.55cm}
\caption{A 2-D synthetic example for adaptive feature acquisition. On the left: data distributed in four clusters. The two features correspond to x and y coordinates, respectively. On the right: accuracy-cost tradeoff curves. Our algorithm can recover the optimal adaptive system whereas a L1-based approach cannot.} \label{fig:synthetic}
\vspace{-.5cm}
\end{wrapfigure}

\noindent
{\bf \textsc{Power of Joint Optimization}:} We return to the problem of prediction under feature budget constrains. 
We illustrate why a simple L1 baseline approach for learning $f_1$ and $g$ would not work using a 2D dataset (Synthetic-2) as shown in Figure \ref{fig:synthetic} (left). The data points are distributed in four clusters, with black triangles and red circles representing two class labels. Let both feature 1 and 2 carry unit acquisition cost. A complex classifier $f_0$ that acquires both features can achieve full accuracy at the cost of 2. 
 In our synthetic example, clusters 1 and 2 are given more data points so that the L1-regularized logistic regression would produce the vertical red dashed line, separating cluster 1 from the others. So feature 1 is acquired for both $g$ and $f_1$. The best such an adaptive system can do is to send cluster 1 to $f_1$ and the other three clusters to the complex classifier $f_0$, incurring an average cost of 1.75, which is sub-optimal.
\textsc{Adapt-Lin}, on the other hand, optimizing between $q,g,f_1$ in an alternating manner, is able to recover the horizontal lines in Figure \ref{fig:synthetic} (left) for $g$ and $f_1$. $g$ sends the first two clusters to the full classifier and the last two clusters to $f_1$. $f_1$ correctly classifies clusters 3 and 4. So all of the examples are correctly classified by the adaptive system; yet only feature 2 needs to be acquired for cluster 3 and 4 so the overall average feature cost is 1.5, as shown by the solid curve in the accuracy-cost tradeoff plot on the right of Figure \ref{fig:synthetic}.  This example shows that the L1 baseline approach is sub-optimal as it doesnot optimize the selection of feature subsets \emph{jointly} for $g$ and $f_1$.

\begin{wraptable}{r}{7cm}
\vspace{-0.2in}
\centering
\caption{Dataset Statistics}
\label{table:datasets}
\resizebox{0.5\columnwidth}{!}{%
\begin{tabular}{@{}cccccc@{}}
\toprule
Dataset   & \#Train & \#Validation & \#Test & \#Features & Feature Costs \\ \midrule
Letters   & 12000   & 4000         & 4000   & 16         & Uniform       \\
MiniBooNE & 45523   & 19510        & 65031  & 50         & Uniform       \\
Forest    & 36603   & 15688        & 58101  & 54         & Uniform       \\
CIFAR10   & 19761   & 8468         & 10000  & 400        & Uniform       \\
Yahoo!    & 141397  & 146769       & 184968 & 519        & CPU units     \\ \bottomrule
\end{tabular}
}
\end{wraptable} 
\noindent
{\bf \textsc{Real Datasets}:} We test various aspects of our algorithms and compare with state-of-the-art feature-budgeted algorithms on five real world benchmark datasets: Letters, MiniBooNE Particle Identification, Forest Covertype datasets from the UCI repository \cite{UCI_repository}, CIFAR-10 \cite{CIFAR10} and Yahoo! Learning to Rank\cite{YahooChallenge2010}. Yahoo! is a ranking dataset where each example is associated with features of a query-document pair together with the relevance rank of the document to the query. There are 519 such features in total; each is associated with an acquisition cost in the set \{1,5,20,50,100,150,200\}, which represents the units of CPU time required to extract the feature and is provided by a Yahoo! employee. The labels are binarized into relevant or not relevant. The task is to learn a model that takes a new query and its associated documents and produce a relevance ranking so that the relevant documents come on top, and to do this using as little feature cost as possible. The performance metric is Average Precision @ 5 following \cite{NanNIPS2016}.
The other datasets have unknown feature costs so we assign costs to be 1 for all features; the aim is to show \textsc{Adapt-Gbrt} successfully selects \emph{sparse} subset of ``usefull'' features for $f_1$ and $g$.
We summarize the statistics of these datasets in Table \ref{table:datasets}. Next, we highlight the key insights from the real dataset experiments. 

\begin{figure*}[htb!]
\centering
\subfigure[MiniBooNE]{\includegraphics[width=.26\linewidth,height=.25\linewidth]{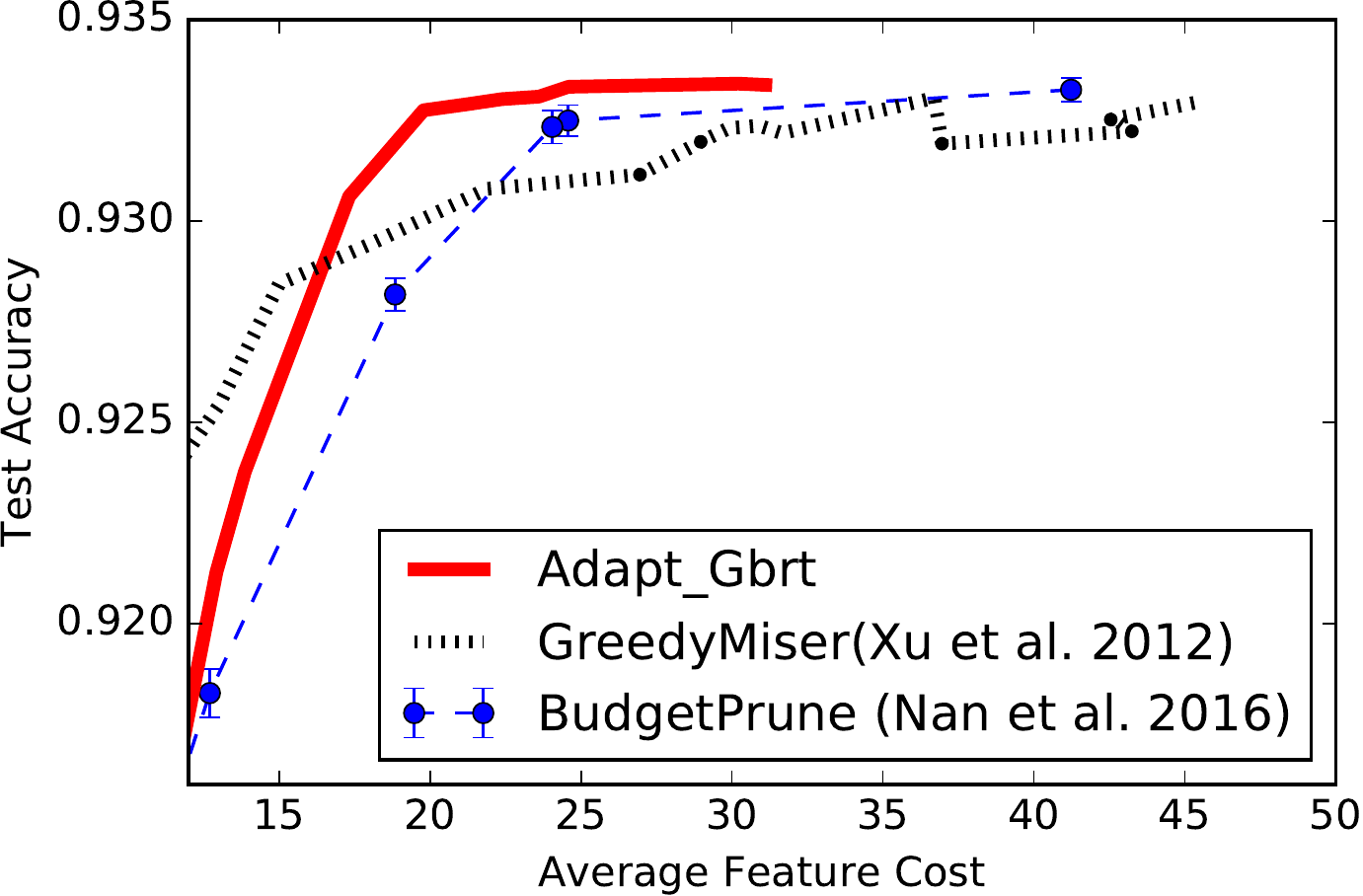}}
\subfigure[Forest Covertype]{\includegraphics[width=.24\linewidth,height=.25\linewidth]{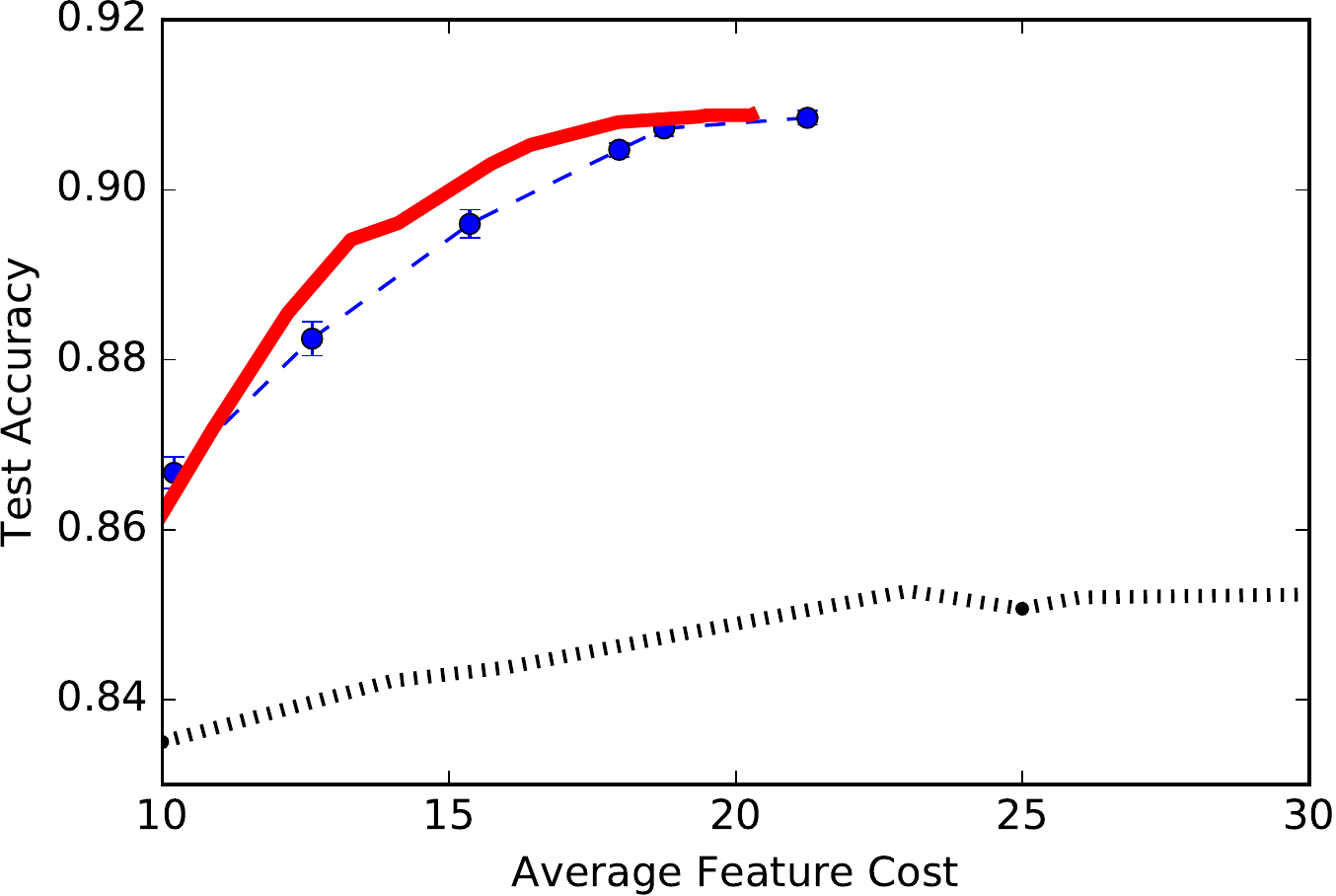}}
\subfigure[Yahoo! Rank]{\includegraphics[width=.24\linewidth,height=.25\linewidth]{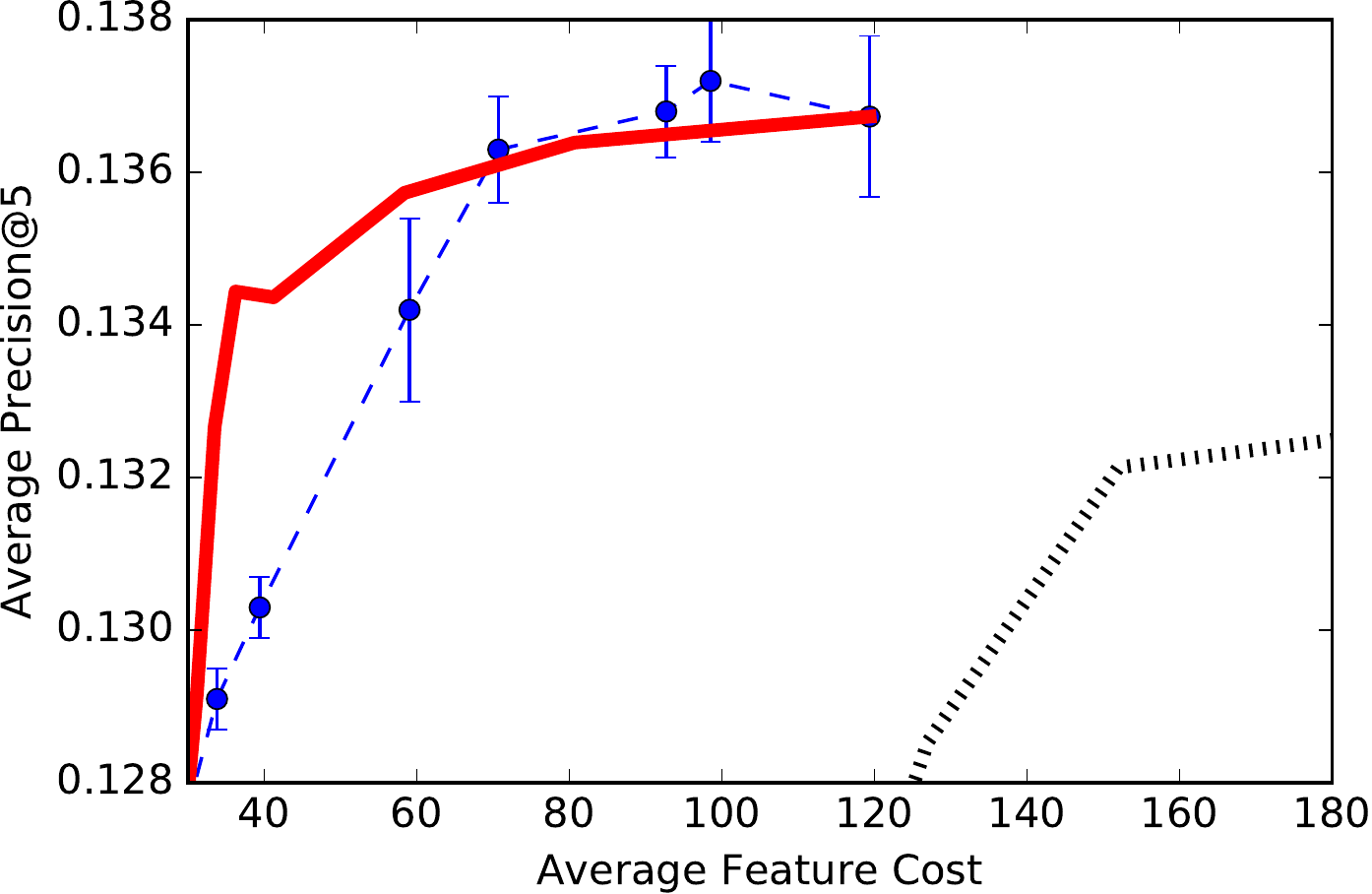}}
\subfigure[CIFAR10]{\includegraphics[width=.24\linewidth,height=.25\linewidth]{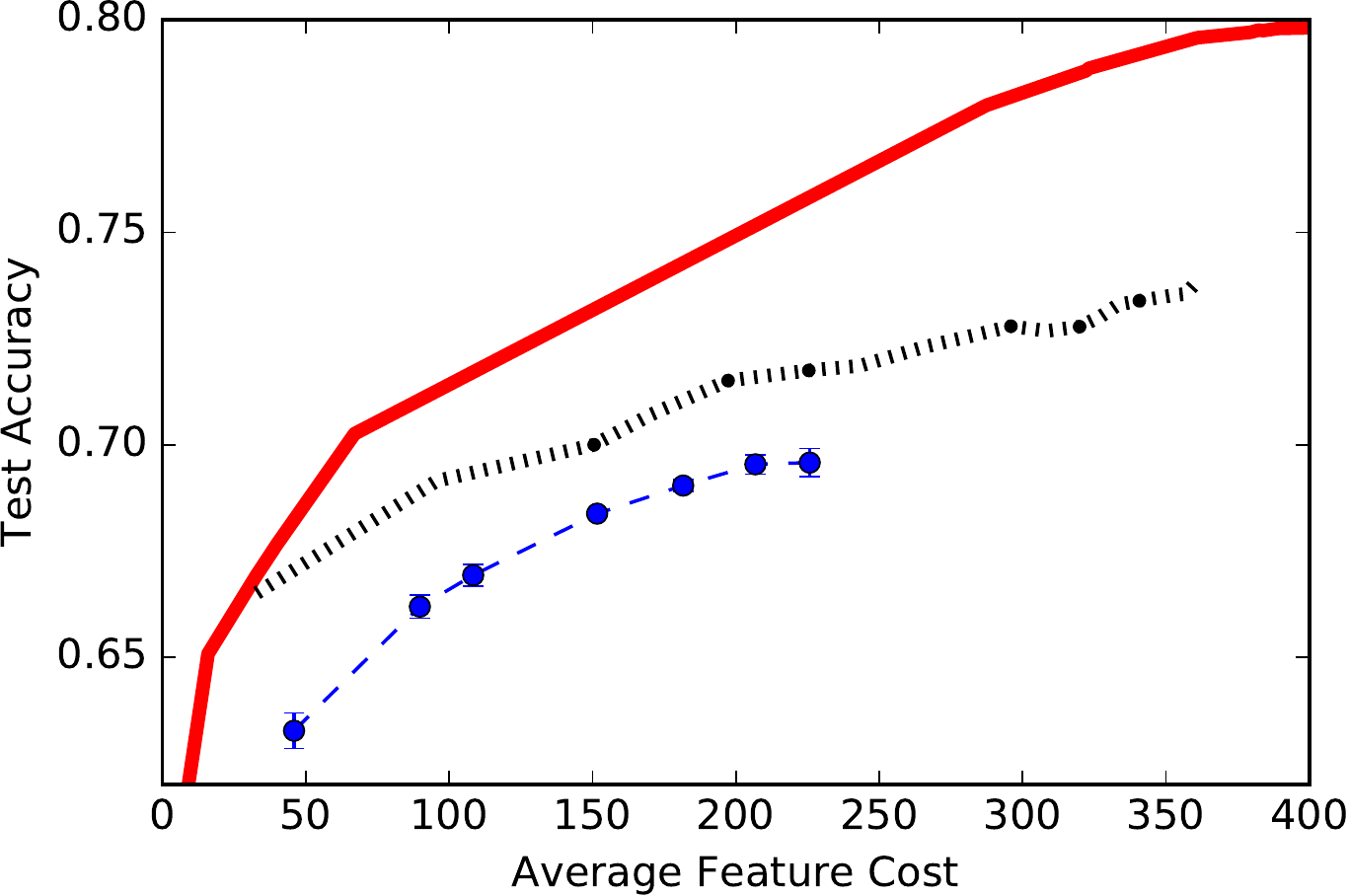}}
\caption{Comparison of \textsc{Adapt-Gbrt} against \textsc{GreedyMiser} and \textsc{BudgetPrune} on four benchmark datasets. RF is used as $f_0$ for \textsc{Adapt-Gbrt} in (a-c) while an RBF-SVM is used as $f_0$ in (d). \textsc{Adapt-Gbrt} achieves better accuracy-cost tradeoff than other methods. The gap is significant in (b) (c) and (d). Note the accuracy of \textsc{GreedyMiser} in (b) never exceeds 0.86 and its precision in (c) slowly rises to 0.138 at cost of 658. We limit the cost range for a clearer comparison.}
\label{fig:experiments}
\end{figure*}

\begin{wrapfigure}{r}{0.37\textwidth}
\centering
\includegraphics[width=0.37\textwidth]{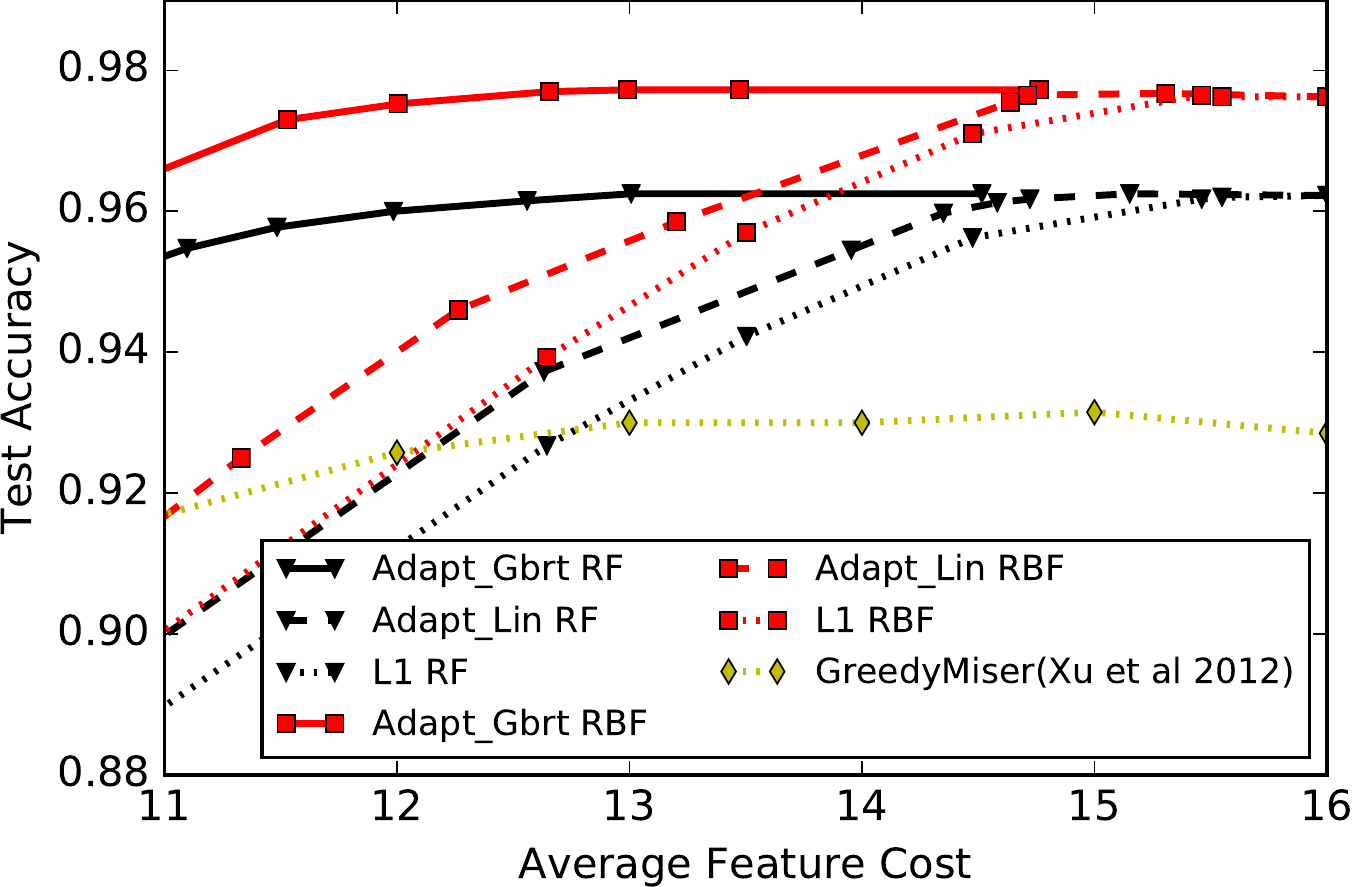}
\vspace{-.3cm}
\caption{Compare the L1 baseline approach, \textsc{Adapt-Lin} and \textsc{Adapt-Gbrt} based on RBF-SVM and RF as $f_0$'s on the Letters dataset.} \label{fig:letters}
\vspace{-.4cm}
\end{wrapfigure}

\noindent
{\bf Generality of Approximation:} Our framework allows approximation of powerful classifiers such as RBF-SVM and Random Forests as shown in Figure \ref{fig:letters} as red and black curves, respectively. In particular, \textsc{Adapt-Gbrt} can well maintain high accuracy while reducing cost. This is a key advantage for our algorithms because we can choose to approximate the $f_0$ that achieves the best accuracy. 
\noindent
{\bf \textsc{Adapt-Lin} Vs L1:} Figure~\ref{fig:letters} shows that \textsc{Adapt-Lin} outperforms L1 baseline method on real dataset as well. Again, this confirms the intuition we have in the Synthetic-2 example as \textsc{Adapt-Lin} is able to iteratively select the common subset of features jointly for $g$ and $f_1$.
\noindent
{\bf \textsc{Adapt-Gbrt} Vs \textsc{Adapt-Lin}:} \textsc{Adapt-Gbrt} leads to significantly better performance than \textsc{Adapt-Lin} in approximating both RBF-SVM and RF as shown in Figure \ref{fig:letters}. This is expected as the non-parametric non-linear classifiers are much more powerful than linear ones.

\noindent
{\bf \textsc{Adapt-Gbrt} Vs \textsc{BudgetPrune}:} Both are bottom-up approaches that benefit from good initializations. In (a), (b) and (c) of Figure \ref{fig:experiments} we let $f_0$ in \textsc{Adapt-Gbrt} be the same RF that \textsc{BudgetPrune} starts with. \textsc{Adapt-Gbrt} is able to maintain high accuracy longer as the budget decreases. Thus, \textsc{Adapt-Gbrt} improves state-of-the-art bottom-up method. Notice in (c) of Figure \ref{fig:experiments} around the cost of 100, \textsc{BudgetPrune} has a spike in precision. We believe this is because the initial pruning improved the generalization performance of RF. But in the cost region of 40-80, \textsc{Adapt-Gbrt} maintains much better accuracy than \textsc{BudgetPrune}. Furthermore, \textsc{Adapt-Gbrt} has the freedom to approximate the best $f_0$ given the problem. So in (d) of Figure \ref{fig:experiments} we see that with $f_0$ being RBF-SVM, \textsc{Adapt-Gbrt} can achieve much higher accuracy than \textsc{BudgetPrune}.

\noindent
{\bf \textsc{Adapt-Gbrt} Vs \textsc{GreedyMiser}:} \textsc{Adapt-Gbrt} outperforms \textsc{GreedyMiser} on all the datasets. The gaps in Figure \ref{fig:letters}, (b) (c) and (d) of Figure \ref{fig:experiments} are especially significant. 

\noindent
{\bf Significant Cost Reduction:} Without sacrificing top accuracies (within 1\%), \textsc{Adapt-Gbrt} reduces average feature costs during test-time by around 63\%, 32\%, 58\%, 12\% and 31\% on MiniBooNE, Forest, Yahoo, Cifar10 and Letters datasets, respectively.

\section{Conclusions}
We presented  an adaptive approximation approach to account for feature acquisition costs that arise in various applications. 
At test-time our method uses a gating function to identify a prediction model among a collection of models that is adapted to the input. The overall goal is to reduce costs without sacrificing accuracy. We learn gating and prediction models by means of a bottom-up strategy that trains low prediction-cost models to approximate high prediction-cost models in regions where low-cost models suffice. On a number of benchmark datasets our method leads to an average of 40\% cost reduction without sacrificing test accuracy (within 1\%). It outperforms state-of-the-art top-down and bottom-up budgeted learning algorithms, with a significant margin in several cases.
%
\subsubsection*{Acknowledgments}

Feng Nan would like to thank Dr Ofer Dekel for ideas and discussions on resource constrained machine learning during an internship in Microsoft Research in summer 2016. Familiarity and intuition gained during the internship contributed to the motivation and formulation in this paper. We also thank Dr Joseph Wang and Tolga Bolukbasi for discussions and helps in experiments.

\bibliography{cost_sensitive_bib}
\bibliographystyle{plain}

\clearpage
\section{Appendix}

\subsection{\textsc{Adapt-Lstsq}}
\paragraph{Other Symmetrized metrics:} KL divergence is not symmetric and leads to widely different properties in terms of approximation. We also consider a symmetrized metric:
$$
D(r(z),s(z)) = \left (\log \frac{r(0)}{r(1)} - \log \frac{s(0)}{s(1)}\right )^2
$$
This metric can be viewed intuitively as a regression of $g(x)=\log(\Pr(1|g;x)/\Pr(0|g;x)$ against the observed log odds ratio of $q(z|x)$.

The main advantage of using KL is that optimizing w.r.t. $q$ can be solved in closed form. The disadvantage we observe is that in some cases, the loss for minimizing w.r.t. $g$, which is a weighted sum of log-losses of opposing directions, becomes quite flat and difficult to optimize especially for linear gating functions. The symmetrized measure, on the other hand, makes the optimization w.r.t. $g$ better conditioned as the gating function $g$ fits directly to the log odds ratio of $q$. However, the disadvantage of using the symmetrized measure is that optimizing w.r.t. $q$ no longer has closed form solution; furthermore, it is even non-convex. We offer an ADMM approach for $q$ optimization.

We still follow an alternating minimization approach. To keep the presentation simply, we assume $g, f_1$ to be linear classifiers and there is no feature costs involved.
To minimize over $q$, we must solve 
\begin{equation}
\begin{array}{rlll}\tag{OPT5}\label{eq:OPT5}
\displaystyle \min_{q_i\in [0,1]} &  \multicolumn{2}{l}{\frac{1}{N} \sum_{i=1}^{N} \left [(1-q_i)A_i+(\log\frac{q_i}{1-q_i}-g(x^{(i)}))^2\right ]} \\
\textrm{s.t.} &  \frac{1}{N} \sum_{i=1}^{N} q_i \leq \text{P}_{\text{full}}, \end{array}
\end{equation}
where $q_i=q(z=0|x^{(i)})$, $A_i=\log(1+e^{-y^{(i)}f_1^Tx^{(i)}})+\log p(y^{(i)}|z^{(i)}=1;f_0)$. Unlike (OPT3), this optimization problem no longer has a closed-form solution. Fortunately, the $q_i$'s in the objective are decoupled and there is only one coupling constraint. We can solve this problem using an ADMM approach \cite{Boyd:2011:DOS:2185815.2185816}.
To optimize over $g$, we simply need to solve a linear least squares problem:
\begin{equation}\tag{OPT6}\label{eq:OPT6}
\displaystyle \min_{g} \frac{1}{N} \sum_{i=1}^{N} (\log\frac{q_i}{1-q_i}-g^T(x^{(i)}))^2.
\end{equation}
To optimize over $f_1$, we solve a weighted logistic regression problem:
\begin{equation}\tag{OPT7}\label{eq:OPT7}
\displaystyle \min_{f_1} \frac{1}{N} \sum_{i=1}^{N} (1-q_i)\log(1+e^{-y^{(i)}f_1^Tx^{(i)}}).
\end{equation}
We shall call the above algorithm \textsc{Adapt-Lstsq}, summarized in Algorithm~\ref{alg:Adapt-lstsq}.
\begin{algorithm}[tb]
	\caption{\textsc{Adapt-Lstsq}}
	\label{alg:Adapt-lstsq}
	\begin{algorithmic}
		\STATE {\bfseries Input:} $(x^{(i)},y^{(i)}),B$
		\STATE Train a full accuracy model $f_0$.
		\STATE Initialize $g, f_1$.
		\REPEAT
		\STATE Solve (OPT5) for $q$ given $g, f_1$. 
		\STATE Solve (OPT6) for $g$ given $q$.
		\STATE Solve (OPT7)for $f_1$ given $q$.
		\UNTIL{convergence}
	\end{algorithmic}
\end{algorithm}

\subsection{Experimental Details}
We provide detailed parameter settings and steps for our experiments here.

\subsection{Synthetic-1 Experiment}
We generate the data in Python using the following command:
\begin{verbatim}
X, y = make_classification(n_samples=1000, flip_y=0.01, n_features=2, 
n_redundant=0, n_informative=2,random_state=17, n_clusters_per_class=2)
\end{verbatim}
For \textsc{Adapt-Gbrt} we used 5 depth-2 trees for $g$ and $f_1$. 

\subsection{Synthetic-2 Experiment:}
We generate 4 clusters on a 2D plane with centers: (1,1), (-1,1), (-1,-1), (-1, -3) and Gaussian noise with standard deviation of 0.01. The first two clusters have 20 examples each and the last two clusters have 15 examples each. We sweep the regularization parameter of L1-regularized logistic regression and recover feature 1 as the sparse subset, which leads to sub-optimal adaptive system.
On the other hand, we can easily train a RBF SVM classifier to correctly classify all clusters and we use it as $f_0$. If we initialize $g$ and $f_1$ with Gaussian distribution centered around 0, \textsc{Adapt-Lin} with can often recover feature 2 as the sparse subset and learn the correct $g$ and $f_1$. Or, we could initialize $g=(1,1)$ and $f_1=(1,1)$ then \textsc{Adapt-Lin} can recover the optimal solution.

\subsection{Letters Dataset \cite{UCI_repository}}
 This letters recognition dataset contains 20000 examples with 16 features, each of which is assigned unit cost. We binarized the labels so that the letters before "N" is class 0 and the letters after and including "N" are class 1.
 We split the examples 12000/4000/4000 for training/validation/test sets. 
 We train RBF SVM and RF (500 trees) with cross-validation as $f_0$. RBF SVM achieves the higher accuracy of 0.978 compared to RF 0.961.
 
 To run the greedy algorithm, we first cross validate L1-regularized logistic regression with 20 C parameters in logspace of [1e-3,1e1]. For each C value, we obtain a classifier and we order the absolute values of its components and threshold them at different levels to recover all 16 possible supports (ranging from 1 feature to all 16 features). We save all such possible supports as we sweep C value. Then for each of the supports we have saved, we train a L2-regularized logistic regression only based on the support features with regularization set to 1 as $f_1$. The gating $g$ is then learned using L2-regularized logistic regression based on the same feature support and pseudo labels of $f_1$ - 1 if it is correctly classified and 0 otherwise. To get different cost-accuracy tradeoff, we sweep the class weights between 0 and 1 so as to influence $g$ to send different fractions of examples to the $f_0$. 
 
 To run \textsc{Adapt-Lin}, we initialize $g$ to be 0 and $f_1$ to be the output of the L2-regularized logistic regression based on all the features. We then perform the alternative minimization for 50 iterations and sweep $\gamma$ between [1e-4,1e0] for 20 points and $\text{P}_\text{full}$ in [0.1,0.9] for 9 points. 
 
 To run \textsc{Adapt-Gbrt}, we use 500 depth 4 trees for $g$ and $f_1$ each. We
 initialize $g$ to be 0 and $f_1$ to be the GreedyMiser output of 500 trees. We then perform the alternative minimization for 30 iterations and sweep $\gamma$ between [1e-1,1e2] for 10 points in logspace and $\text{P}_\text{full}$ in [0.1,0.9] for 9 points. In addition, we also sweep the learning rate for GBRT for 9 points between [0.1,1].
 
 For fair comparison, we run \textsc{GreedyMiser} with 1000 depth 4 trees so that the model size matches that of \textsc{Adapt-Gbrt}. The learning rate is swept between [1e-5,1] with 20 points and the $\lambda$ is swept between [0.1, 100] with 20 points.
 
 Finally, we evaluate all the resulting systems from the parameter sweeps of all the algorithms on validation data and choose the efficient frontier and use the corresponding settings to evaluate and plot the test performance. 
 
 \subsection{MiniBooNE Particle Identification and Forest Covertype Datasets \cite{UCI_repository}:}
 The MiniBooNE data set is a binary classification task to distinguish electron neutrinos from muon neutrinos. There are $45523/19510/65031$ examples in training/validation/test sets. Each example has 50 features, each with unit cost. 
 The Forest data set contains cartographic variables to predict 7 forest cover types. There are $36603/15688/58101$ examples in training/validation/test sets. Each example has 54 features, each with unit cost.

 We use the unpruned RF of \textsc{BudgetPrune} \cite{NanNIPS2016} as $f_0$ (40 trees for both datasets.) The settings for \textsc{Adapt-Gbrt} are the following.
 For MiniBooNE we use 100 depth 4 trees for $g$ and $f_1$ each. We
  initialize $g$ to be 0 and $f_1$ to be the GreedyMiser output of 100 trees. We then perform the alternative minimization for 50 iterations and sweep $\gamma$ between [1e-1,1e2] for 20 points in logspace and $\text{P}_\text{full}$ in [0.1,0.9] for 9 points. In addition, we also sweep the learning rate for GBRT for 9 points between [0.1,1].
  For Forest we use 500 depth 4 trees for $g$ and $f_1$ each. We
    initialize $g$ to be 0 and $f_1$ to be the GreedyMiser output of 500 trees. We then perform the alternative minimization for 50 iterations and sweep $\gamma$ between [1e-1,1e2] for 20 points in logspace and $\text{P}_\text{full}$ in [0.1,0.9] for 9 points. In addition, we also sweep the learning rate for GBRT for 9 points between [0.1,1].
    
For fair comparison, we run \textsc{GreedyMiser} with 200 depth 4 trees so that the model size matches that of \textsc{Adapt-Gbrt} for MiniBooNE. We run \textsc{GreedyMiser} with 1000 depth 4 trees so that the model size matches that of \textsc{Adapt-Gbrt} for Forest.

   Finally, we evaluate all the resulting systems from the parameter sweeps on validation data and choose the efficient frontier and use the corresponding settings to evaluate and plot the test performance. 
 
 \subsection{Yahoo! Learning to Rank\cite{YahooChallenge2010}:}
  This ranking dataset consists of 473134 web documents and 19944 queries. Each example is associated with features of a query-document pair together with the relevance rank of the document to the query. There are 519 such features in total; each is associated with an acquisition cost in the set \{1,5,20,50,100,150,200\}, which represents the units of CPU time required to extract the feature and is provided by a Yahoo! employee. The labels are binarized into relevant or not relevant. The task is to learn a model that takes a new query and its associated documents and produce a relevance ranking so that the relevant documents come on top, and to do this using as little feature cost as possible. The performance metric is Average Precision @ 5 following \cite{NanNIPS2016}. 

   We use the unpruned RF of \textsc{BudgetPrune} \cite{NanNIPS2016} as $f_0$ (140 trees for both datasets.) The settings for \textsc{Adapt-Gbrt} are the following. we use 100 depth 4 trees for $g$ and $f_1$ each. We
       initialize $g$ to be 0 and $f_1$ to be the \textsc{GreedyMiser} output of 100 trees. We then perform the alternative minimization for 20 iterations and sweep $\gamma$ between [1e-1,1e3] for 30 points in logspace and $\text{P}_\text{full}$ in [0.1,0.9] for 9 points. In addition, we also sweep the learning rate for GBRT for 9 points between [0.1,1].

For fair comparison, we run \textsc{GreedyMiser} with 200 depth 4 trees so that the model size matches that of \textsc{Adapt-Gbrt} for Yahoo. 
  
  Finally, we evaluate all the resulting systems from the parameter sweeps on validation data and choose the efficient frontier and use the corresponding settings to evaluate and plot the test performance. 
  
 \subsection{CIFAR10 \cite{CIFAR10}:}
 CIFAR-10 data set consists of 32x32 colour images in 10 classes.
 400 features for each image are extracted using technique described in \cite{ICML2011Coates_485}. 
 The data are binarized by combining the first 5 classes into one class and the others into the second class. There are $19,761/8,468/10,000$ examples in training/validation/test sets.
 \textsc{BudgetPrune} starts with a RF of 40 trees, which achieves an accuracy of 69\%. 
     We use an RBF-SVM as $f_0$ that achieves a test accuracy of 79.5\%. The settings for \textsc{Adapt-Gbrt} are the following. we use 200 depth 5 trees for $g$ and $f_1$ each. We
         initialize $g$ to be 0 and $f_1$ to be the \textsc{GreedyMiser} output of 200 trees. We then perform the alternative minimization for 50 iterations and sweep $\gamma$ between [1e-4,10] for 15 points in logspace and $\text{P}_\text{full}$ in [0.1,0.9] for 9 points. In addition, we also sweep the learning rate for GBRT for 10 points between [0.01,1].
  
  For fair comparison, we run \textsc{GreedyMiser} with 400 depth 5 trees so that the model size matches that of \textsc{Adapt-Gbrt}. 
    
    Finally, we evaluate all the resulting systems from the parameter sweeps on validation data and choose the efficient frontier and use the corresponding settings to evaluate and plot the test performance.

\end{document}